\documentclass[sn-basic,Numbered]{sn-jnl}

\usepackage{graphicx}%
\usepackage{multirow}%
\usepackage{amsmath,amssymb,amsfonts}%
\usepackage{amsthm}%
\usepackage{mathrsfs}%
\usepackage[title]{appendix}%
\usepackage{xcolor}%
\usepackage{textcomp}%
\usepackage{manyfoot}%
\usepackage{booktabs}%
\usepackage{algorithm}%
\usepackage{algorithmicx}%
\usepackage{algpseudocode}%
\usepackage{listings}%
\usepackage{caption}
\usepackage{subcaption}

\theoremstyle{thmstyleone}%
\theoremstyle{thmstyletwo}%

\theoremstyle{thmstylethree}%

\begin{document}

\title{Virtual Intraoperative CT (viCT): Sequential Anatomic Updates for Modeling Tissue Resection Throughout Endoscopic Sinus Surgery}


% --- Authors (Springer "sn-article"/AOFM-style) ---
\author[1]{\fnm{Nicole M.} \sur{Gunderson}}
\author[2]{\fnm{Graham J.} \sur{Harris}}
\author[2]{\fnm{Jeremy S.} \sur{Ruthberg}}
\author[1]{\fnm{Pengcheng} \sur{Chen}}
\author[1]{\fnm{Di} \sur{Mao}}
\author[2]{\fnm{Randall A.} \sur{Bly}}
\author[2]{\fnm{Waleed M.} \sur{Abuzeid}}
\author*[1]{\fnm{Eric J.} \sur{Seibel}}\email{eseibel@uw.edu}

% If Nicole is also corresponding, uncomment and add her email:
% \author*[1]{\fnm{Nicole M.} \sur{Gunderson}}\email{<nicole_email_here>}

% --- Affiliations ---
\affil[1]{\orgdiv{Department of Mechanical Engineering}, \orgname{University of Washington},
\orgaddress{\city{Seattle}, \state{WA}, \country{USA}}}

\affil[2]{\orgdiv{Department of Otolaryngology--Head and Neck Surgery, University of Washington Medical Center},
\orgname{University of Washington},
\orgaddress{\city{Seattle}, \state{WA}, \country{USA}}}

%%================================%%
%% Sample for structured abstract %%
%%================================%%



\maketitle

\begin{abstract}

\textbf{Purpose:} Incomplete dissection is a common cause of persistent disease and revision endoscopic sinus surgery (ESS) in chronic rhinosinusitis. Current image-guided surgery systems typically reference static preoperative CT (pCT), and do not model evolving resection boundaries. We present Virtual Intraoperative CT (viCT), a method for sequentially updating pCT throughout ESS using intraoperative 3D reconstructions from monocular endoscopic video to enable visualization of evolving anatomy in CT format.

\textbf{Methods:} Monocular endoscopic video is processed using a depth-supervised NeRF framework with virtual stereo synthesis to generate metrically scaled 3D reconstructions at multiple surgical intervals. Reconstructions undergo rigid, landmark-based registration in 3D Slicer guided by anatomical correspondences, and are then voxelized into the pCT grid. viCT volumes were generated using a ray-based voxel occupancy comparison between pCT and reconstruction to delete outdated voxels and remap preserved anatomy and updated boundaries. Performance is evaluated in a cadaveric feasibility study of four specimens across four ESS stages using volumetric overlap (DSC, Jaccard) and surface metrics (HD95, Chamfer, MSD, RMSD), alongside qualitative slice comparisons to ground-truth interval CT.

\textbf{Results:} viCT updates show close agreement with ground-truth anatomy across surgical stages, with submillimeter mean surface errors. Dice Similarity Coefficient (DSC) = 0.88 ± 0.05 and Jaccard Index = 0.79 ± 0.07, and Hausdorff Distance 95\% (HD95) = 0.69 ± 0.28 mm, Chamfer Distance = 0.09 ± 0.05 mm, Mean Surface Distance (MSD) = 0.11 ± 0.05 mm, and Root Mean Square Distance (RMSD) = 0.32 ± 0.10 mm.

\textbf{Conclusion:} viCT enables CT-format, sequential anatomic updating in an ESS setting without ancillary hardware. Future work will focus on fully automating registration, expanding validation in live cases, and optimizing runtime for real-time deployment.

\end{abstract}

\keywords{Virtual intraoperative CT, Endoscopic sinus surgery, Monocular endoscopy, NeRF, CT updating, Surgical navigation}

\section{Introduction}\label{sec1}

\subsection{Chronic Rhinosinusitis and Endoscopic Sinus Surgery}

Chronic rhinosinusitis (CRS) is a persistent inflammatory condition of the nasal cavity and paranasal sinuses characterized by nasal obstruction, drainage, facial pressure, and olfactory dysfunction. CRS affects approximately 1 in 8 adults in the United States and accounts for tens of millions of clinical diagnoses annually. \cite{Rosenfeld2015SinusitisGuideline} The disease substantially impairs health-related quality-of-life, with a mean utility score of 0.65 (0 = death, 1 = perfect health), comparable to other major chronic illnesses. \cite{Soler2011HealthUtility, Remenschneider2015EQ5D}

Endoscopic sinus surgery (ESS) is the primary treatment for CRS refractory to medical therapy. Approximately 28.3\% of CRS patients undergo ESS, with 75\% achieving sustained symptom relief. \cite{Benson2022CRSwNPBurden} However, revision surgery is required in 15–20\% of cases, \cite{Smith2019LongTermESS, Hopkins2009NasalPolyposisAudit, Smith2019RevisionRates, Senior1998FESSOutcomes} and among patients with multiple procedures, 29\% undergo a third surgery within 36 months. Revision ESS is associated with lower success rates and independently predicts poorer outcomes. \cite{Benson2022CRSwNPBurden}

Incomplete dissection is a leading cause of surgical failure. \cite{Benson2022CRSwNPBurden} Residual bony partitions and persistent inflammation have been identified in 64–96\% of postoperative CT evaluations. \cite{Musy2004RevisionAnatomy, Ramadan1999SurgicalFailure, Okushi2012ResidualEthmoid, Gore2013CentralSinusRevision, Khalil2011RadiologicalRevision, Baban2020RadiologicalRevision} In a prospective study, intraoperative CT performed at the conclusion of ESS altered the surgical plan in 30\% of patients by prompting additional tissue removal. \cite{Jackman2008IntraoperativeCT} Collectively, these findings demonstrate that incomplete primary ESS contributes to poorer clinical outcomes and that updated imaging can meaningfully influence intraoperative decision-making.

\subsection{Existing IGS Systems}

To mitigate incomplete resection, commercial image-guided surgery (IGS) platforms—including Stryker’s Scopis, GE’s InstaTrak, Brainlab’s Curve, and the Medtronic Fusion system—provide real-time visualization of surgical instruments or endoscope tips relative to preoperative CT (pCT) images. \cite{Douglas2022FastAnatomicMapping, Juvekar2023ToolTipTracking, Schmale2021IGSStateOfTheArt}

However, these systems do not update pCT data to reflect evolving tissue removal. Surgeons must manually infer resection boundaries, increasing operative time and susceptibility to registration drift. Reported tracking errors can exceed 2 mm—clinically significant given the narrow sinonasal corridors—and may contribute to incomplete resection. \cite{Lorenz2006KolibriNavigation, Citardi2017NextGenNavigation, Isikay2024Narrative3DVisualization}

\subsection{Intraoperative CT}

Intraoperative CT (iCT) remains the most reliable method for assessing extent of resection during ESS. Despite its accuracy, iCT is rarely employed due to workflow interruptions of 30–40 minutes, prolonged anesthesia time, increased cost, and radiation exposure. \cite{Jackman2008IntraoperativeCT}

There remains a need for a method that approximates the anatomical fidelity of iCT while preserving operative efficiency. An approach that generates updated, quantitative 3D CT-based models without added radiation or surgical delay could provide intuitive visualization of evolving resection boundaries and improve intraoperative assessment of tissue removal.

\subsection{Contributions}

This work introduces Virtual Intraoperative CT (viCT), a framework that produces sequential, CT-format anatomic updates during endoscopic sinus surgery from monocular endoscopic video. Our key contributions are:
\begin{itemize}
    \item A pipeline that maps metrically scaled, NeRF-based intraoperative 3D reconstructions into the native preoperative CT (pCT) grid to enable direct axial/coronal/sagittal viCT visualization without ancillary hardware.
    \item A ray-based volumetric mask construction and voxel-level update rule that deletes pCT regions corresponding to resected tissue while preserving unchanged anatomy in pCT coordinates.
    \item A multi-stage cadaveric feasibility validation (four specimens; four ESS stages with interval CT ground truth) demonstrating high volumetric overlap and submillimeter surface agreement between viCT and postoperative CT across surgical progression.
\end{itemize}

\section{Methods}\label{sec2}

\subsection{3D Reconstruction}

Metrically scaled intraoperative geometry was generated using a depth-supervised Neural Radiance Field (NeRF) framework shown in Figure 1, with input to this algorithm shown in Figure A1. Sequential monocular endoscopic frames (4 mm rigid scope) were converted into a spatially consistent 3D representation suitable for CT-grid voxelization.

\paragraph{Camera Pose Estimation}
Frames were processed with COLMAP to estimate intrinsics $\mathbf{K}$ and poses $\mathbf{T}_i=[\mathbf{R}_i|\mathbf{t}_i]\in SE(3)$, with projection
\begin{equation}
\mathbf{x}_i \sim \mathbf{K}(\mathbf{R}_i\mathbf{X}+\mathbf{t}_i).
\end{equation}

\paragraph{NeRF Initialization (Hash + Spherical Harmonics)}
NeRF models radiance and density as
\begin{equation}
F_{\theta}:(\mathbf{x},\mathbf{d})\rightarrow (\sigma,\mathbf{c}),
\end{equation}
where $\mathbf{x}\in\mathbb{R}^3$ is position and $\mathbf{d}$ is viewing direction. Following EndoPerfect, positions are encoded with multi-resolution hash encoding $\phi(\mathbf{x})$, and view-dependence is represented with spherical harmonics $\gamma(\mathbf{d})$:
\begin{equation}
\gamma(\mathbf{d})=\big[ Y_0^0(\mathbf{d}),\,Y_1^{-1}(\mathbf{d}),\ldots, Y_{\ell_{\max}}^{\ell_{\max}}(\mathbf{d}) \big].
\end{equation}
A coarse-to-fine NeRF is trained by minimizing image reconstruction error
\begin{equation}
\mathcal{L}_{\text{img}}=\sum_{i}\|I_i-\hat{I}_i\|_2^2,
\end{equation}
where $\hat{I}_i$ is rendered via volume rendering along camera rays.

\paragraph{NeRF Rendering}
For a ray $\mathbf{r}(t)=\mathbf{o}+t\mathbf{d}$ with samples $\{t_k\}_{k=1}^N$, EndoPerfect uses the standard discrete volume rendering form
\begin{equation}
\hat{\mathbf{C}}(\mathbf{r})=\sum_{k=1}^{N} T_k\big(1-e^{-\sigma_k\delta_k}\big)\,\mathbf{c}_k,
\qquad
T_k=\exp\!\left(-\sum_{j<k}\sigma_j\delta_j\right),
\end{equation}
where $\sigma_k=\sigma(\mathbf{r}(t_k))$, $\mathbf{c}_k=\mathbf{c}(\mathbf{r}(t_k),\mathbf{d})$, and $\delta_k=t_{k+1}-t_k$.

\paragraph{Optimized Novel Stereo View Generation}
To obtain metric depth without a physical stereo endoscope, EndoPerfect synthesizes a novel stereo partner view inside the trained NeRF. For each training view with pose $\mathbf{T}_{\text{train}}$, a novel pose $\mathbf{T}_{\text{novel}}$ is optimized to maximize stereo matchability using zero-mean normalized cross-correlation (ZNCC) while enforcing geometric constraints (coplanarity, fixed baseline $b$, and consistent orientation):
\begin{equation}
\mathbf{T}_{\text{novel}}^{*}
=
\arg\max_{\mathbf{T}_{\text{novel}}\in \mathcal{C}}
\ \mathrm{ZNCC}\!\left(I_{\text{train}},\, I_{\text{novel}}(\mathbf{T}_{\text{novel}})\right),
\end{equation}
with constraint set $\mathcal{C}$ enforcing (i) a fixed baseline $\| \mathbf{p}_{\text{novel}}-\mathbf{p}_{\text{train}}\|=b$, (ii) coplanar motion relative to the training camera, and (iii) matched viewing direction for robust correspondence. Rendered pairs $(I_{\text{train}}, I_{\text{novel}})$ are then used for stereo depth estimation.

\paragraph{Stereo Depth and Metric Scale}
Given disparity $D(\cdot)$ predicted from the rendered stereo pair, depth is recovered by binocular triangulation:
\begin{equation}
d(u,v)=D\!\left(I_{\text{train}}, I_{\text{novel}}\right)(u,v),
\qquad
Z(u,v)=\frac{fb}{d(u,v)},
\end{equation}
where $f$ is focal length and $b$ is the enforced baseline.

\paragraph{Depth-Supervised Iterative Refinement}
Depth maps supervise subsequent NeRF training rounds (depth-supervised NeRF/DS-NeRF style), yielding an overall objective
\begin{equation}
\mathcal{L}=\mathcal{L}_{\text{img}}+\lambda_d\,\mathcal{L}_{\text{depth}}.
\end{equation}
This loop (NeRF $\rightarrow$ optimized stereo views $\rightarrow$ stereo depth $\rightarrow$ depth-supervised NeRF) is repeated until depth convergence between iterations, after which final depth maps are fused across views to produce a dense, metrically scaled reconstruction. The resulting 3D model is then rigidly registered to the preoperative CT (pCT) for viCT updating. \cite{Chen2024HybridNeRFEndoscopy, Gunderson2025HighFidelityESS, Ruthberg2025VisionGuidedNavigation}

\begin{figure}[htp]
    \centering
    \includegraphics[width=0.94\linewidth]{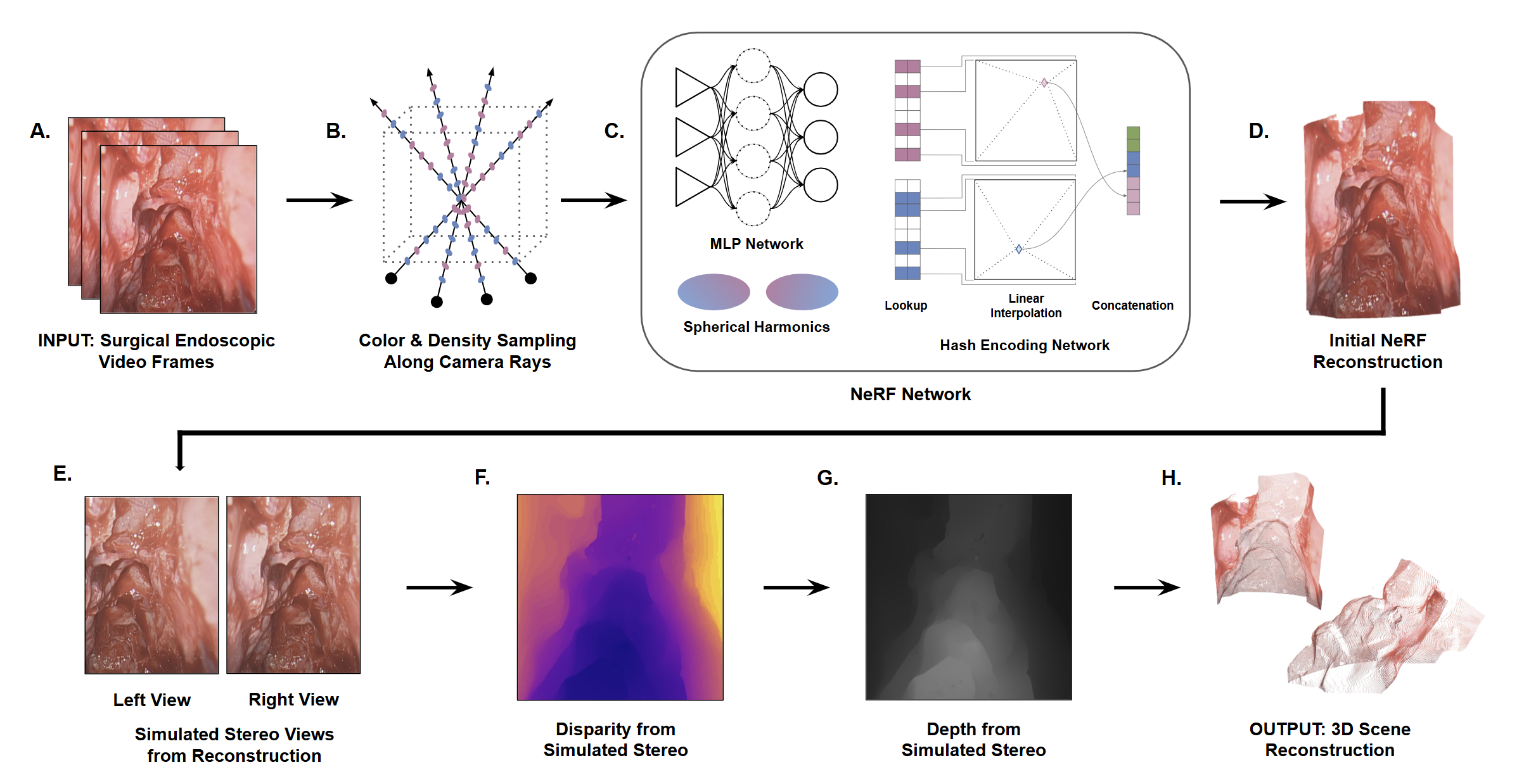} 
    \caption{Flowchart of simulated stereoscopy NeRF-based reconstruction workflow.}
    \label{figure1}
\end{figure}

\subsection{Registration}

To generate viCT volumes, each intraoperative 3D reconstruction was rigidly registered to the corresponding preoperative CT (pCT). Semi-automatic, landmark-based rigid registration was performed in 3D Slicer using anatomically corresponding fiducials visible in both datasets. Landmarks were selected from stable bony structures, placed by researchers, and verified by an otolaryngologist to ensure anatomical consistency. 

Between 3–10 landmarks were used per specimen and surgical interval. Common fiducials included the nasofrontal beak, middle turbinate axilla, junction of the maxillary sinus roof with the lamina papyracea in the plane of the posterior maxillary wall, sphenoid sinus landmarks (ostium, rostrum, floor, posterior wall), orbital entry of the anterior ethmoid artery, cribriform lateral lamella/fovea ethmoidalis junction, and basal lamella/skull base attachment.

\subsection{Virtual Intraoperative CT (viCT) Updating}

\begin{figure}[htp]
    \centering
    \includegraphics[width=0.95\linewidth]{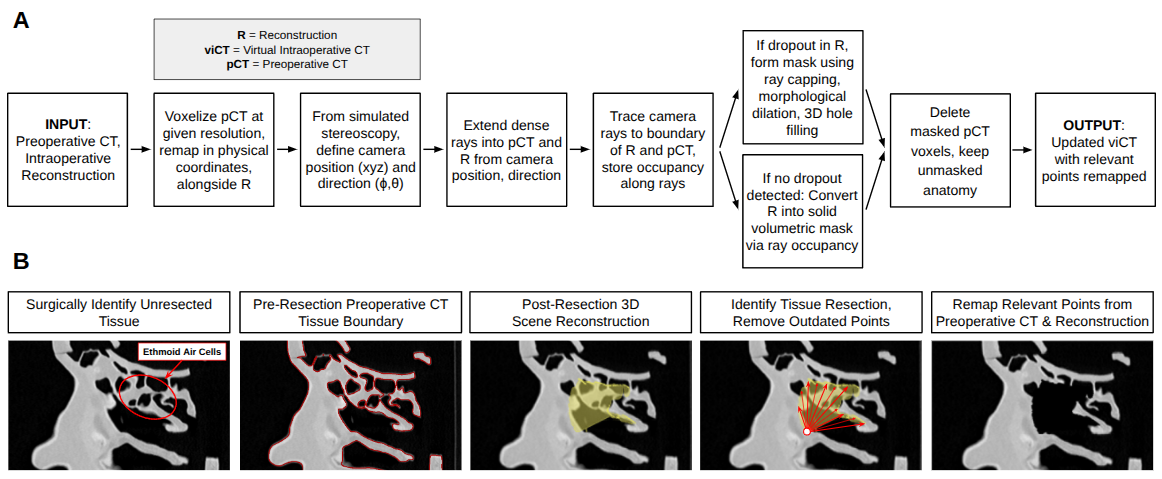} 
    \caption{Ray-based viCT updating workflow and example of tissue resection modeling.}
    \label{figure2}
\end{figure}

The viCT update is computed directly in the \emph{native preoperative CT (pCT) voxel grid} using DICOM metadata (origin, spacing, orientation). Let the pCT be
\[
I_{\mathrm{pCT}} : \Omega \subset \mathbb{Z}^3 \rightarrow \mathbb{R},
\]
with voxel index $\mathbf{i}=(i_x,i_y,i_z)$. The corresponding physical coordinate $\mathbf{x}\in\mathbb{R}^3$ is

\begin{equation}
\mathbf{x}(\mathbf{i})
=
\mathbf{o}_{\mathrm{CT}}
+
\mathbf{D}_{\mathrm{CT}}
\begin{bmatrix}
s_x i_x\\
s_y i_y\\
s_z i_z
\end{bmatrix},
\label{eq:dicom_mapping}
\end{equation}

where $\mathbf{o}_{\mathrm{CT}}$ is the DICOM origin, $\mathbf{D}_{\mathrm{CT}} \in SO(3)$ the direction cosine matrix, and $(s_x,s_y,s_z)$ the voxel spacings (mm). All reconstruction-derived geometry is rigidly registered and expressed in this same physical coordinate frame.

\paragraph{Binary anatomical mask in CT space}

A binary anatomy mask is first generated from the pCT via thresholding:

\begin{equation}
M_{\mathrm{pCT}}(\mathbf{i})
=
\mathbb{I}\!\left(I_{\mathrm{pCT}}(\mathbf{i}) > \tau\right),
\label{eq:pct_mask}
\end{equation}

where $\tau$ is a fixed Hounsfield Unit (HU) threshold (e.g., $\tau=-300$ HU). Importantly, the full HU-valued pCT volume $I_{\mathrm{pCT}}$ is preserved throughout the update process; only voxel occupancy is modified.

\paragraph{Volumetric reconstruction mask from STL + FCSV}

Intraoperative reconstructions are exported as a surface mesh (STL) with an associated fiducial point (FCSV) representing the reconstructed camera origin $\mathbf{o}_c$ in physical space. Because STL meshes may be hollow or incomplete, a volumetric occupancy mask is constructed via ray-based voxelization.

Let $\{\mathbf{v}_m\}_{m=1}^{M}$ denote mesh vertices. For each vertex,

\begin{equation}
\mathbf{d}_m = \frac{\mathbf{v}_m - \mathbf{o}_c}{\|\mathbf{v}_m - \mathbf{o}_c\|},
\qquad
\ell_m = \|\mathbf{v}_m - \mathbf{o}_c\|.
\end{equation}

Rays are parameterized as

\begin{equation}
\mathbf{r}_m(t) = \mathbf{o}_c + t \mathbf{d}_m,
\qquad t \in [0,\ell_m].
\label{eq:ray_param}
\end{equation}

Sampled points along each ray are mapped to voxel indices using the inverse of Eq.~\eqref{eq:dicom_mapping}, generating a sparse occupancy scaffold. Binary dilation, morphological closing, and hole filling are then applied to produce a solid, watertight volumetric mask

\[
M_{\mathrm{rec}} : \Omega \rightarrow \{0,1\},
\]

expressed in the pCT grid.

\paragraph{Voxel-wise CT update rule}

Resected regions are defined as voxels present in the pCT mask but not supported by the intraoperative reconstruction. The updated occupancy mask is

\begin{equation}
M_{\mathrm{viCT}}
=
M_{\mathrm{pCT}} \wedge \neg M_{\mathrm{rec}}.
\label{eq:vict_mask}
\end{equation}

The final viCT image is then constructed by modifying the original HU-valued pCT volume:

\begin{equation}
I_{\mathrm{viCT}}(\mathbf{i})
=
\begin{cases}
I_{\mathrm{pCT}}(\mathbf{i}), & M_{\mathrm{viCT}}(\mathbf{i}) = 1, \\
I_{\mathrm{air}}, & \text{otherwise},
\end{cases}
\label{eq:vict_intensity}
\end{equation}

where $I_{\mathrm{air}}$ corresponds to an air-equivalent HU value (e.g., $-1000$ HU). Thus, unchanged anatomy retains its original HU intensities and DICOM metadata, while resected tissue is removed by intensity reassignment. The output viCT is therefore a full HU-valued DICOM volume in the native pCT coordinate system.

\paragraph{Physical distance computation}

For quantitative validation against interval CT, binary masks derived from $I_{\mathrm{viCT}}$ and ground-truth CT are compared within a reconstruction-defined ROI. Surface distances computed in voxel units are converted to millimeters using mean voxel spacing $\bar{s}=\tfrac{1}{3}(s_x+s_y+s_z)$:

\begin{equation}
d_{\mathrm{mm}} = d_{\mathrm{vox}} \, \bar{s}.
\end{equation}

\noindent
This updating process may be repeated iteratively as new intraoperative reconstructions are generated, producing sequential HU-preserving CT-format updates throughout surgery as shown in Figure 2.

\section{Experiment}\label{sec3}

\subsection{Surgical Method}

Four cadaver heads underwent CT imaging at 4 timepoints to establish ground truth of the updated intraoperative anatomy: (1) pre-operatively, (2) after execution of maxillary antrostomy, (3) after partial  anterior ethmoidectomy, and (4) postoperatively after posterior ethmoidectomy and sphenoidotomy. An otolaryngology surgeon captured monocular endoscopic video of the sinus at all relevant angles and locations to display tissue change and removal as surgery progressed with each interval. The video of the surgery was analyzed after the ESS using 3D scene reconstructions for each surgical interval, along with viCT volumes. 

\section{Results}\label{sec4}

\subsection{Qualitative Data}

The qualitative comparisons between each viCT update and its corresponding postoperative ground-truth CT further highlight the fidelity of the reconstructed anatomy. Figure 3 and Supplementary Figures 6–8 show side-by-side visualizations of axial, coronal, and sagittal slices are shown for sequential surgical intervals for cadaveric specimens 1, 2, 3, and 4, respectively.

\begin{figure}[p]
    \centering
    \begin{subfigure}{0.9\linewidth}
        \centering
        \includegraphics[width=\linewidth]{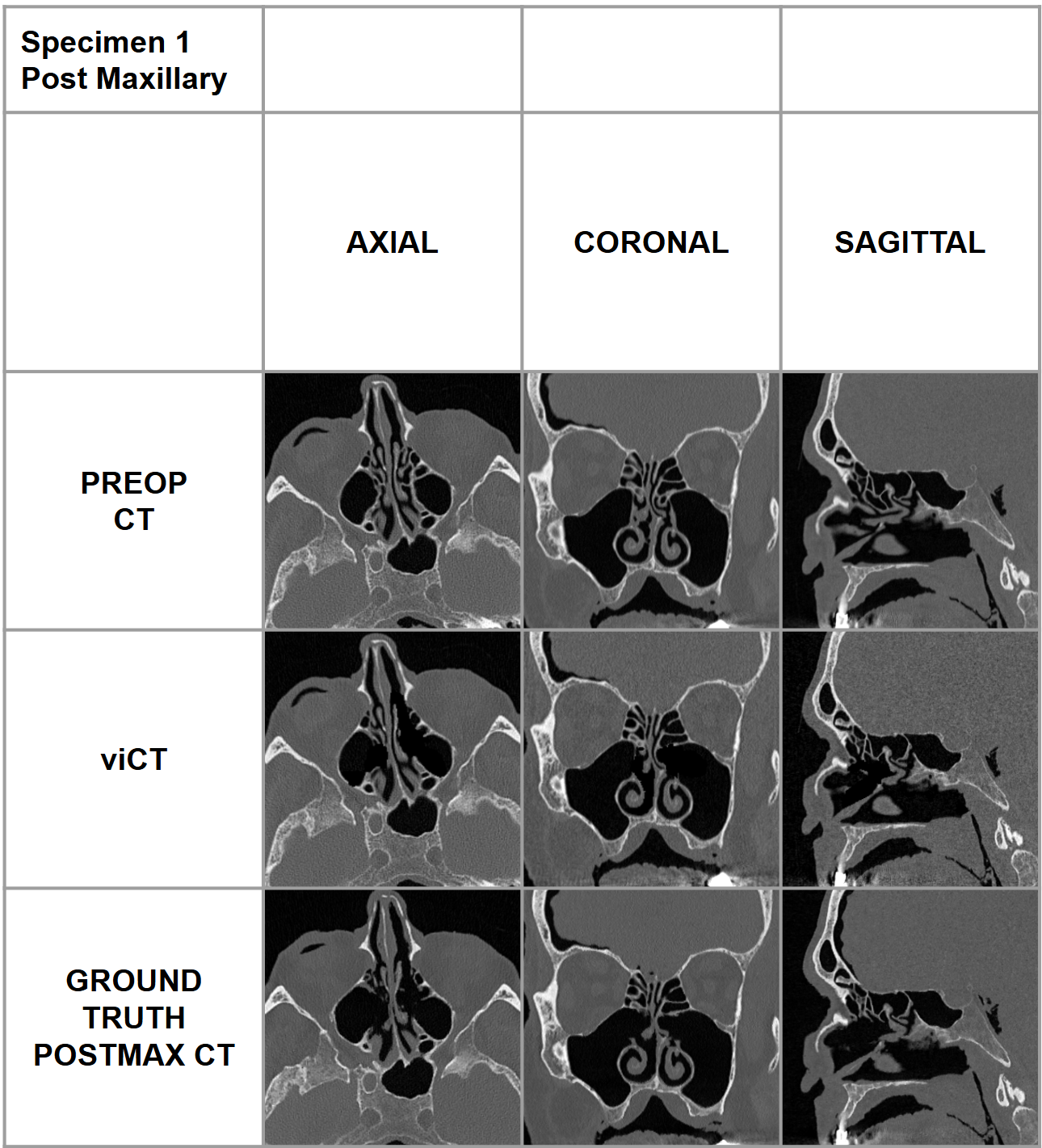}
        \caption{Post Maxillary surgical interval}
        \label{fig:fig4A}
    \end{subfigure}
\end{figure}

\begin{figure}[p]\ContinuedFloat
    \centering
    \begin{subfigure}{0.9\linewidth}
        \centering
        \includegraphics[width=\linewidth]{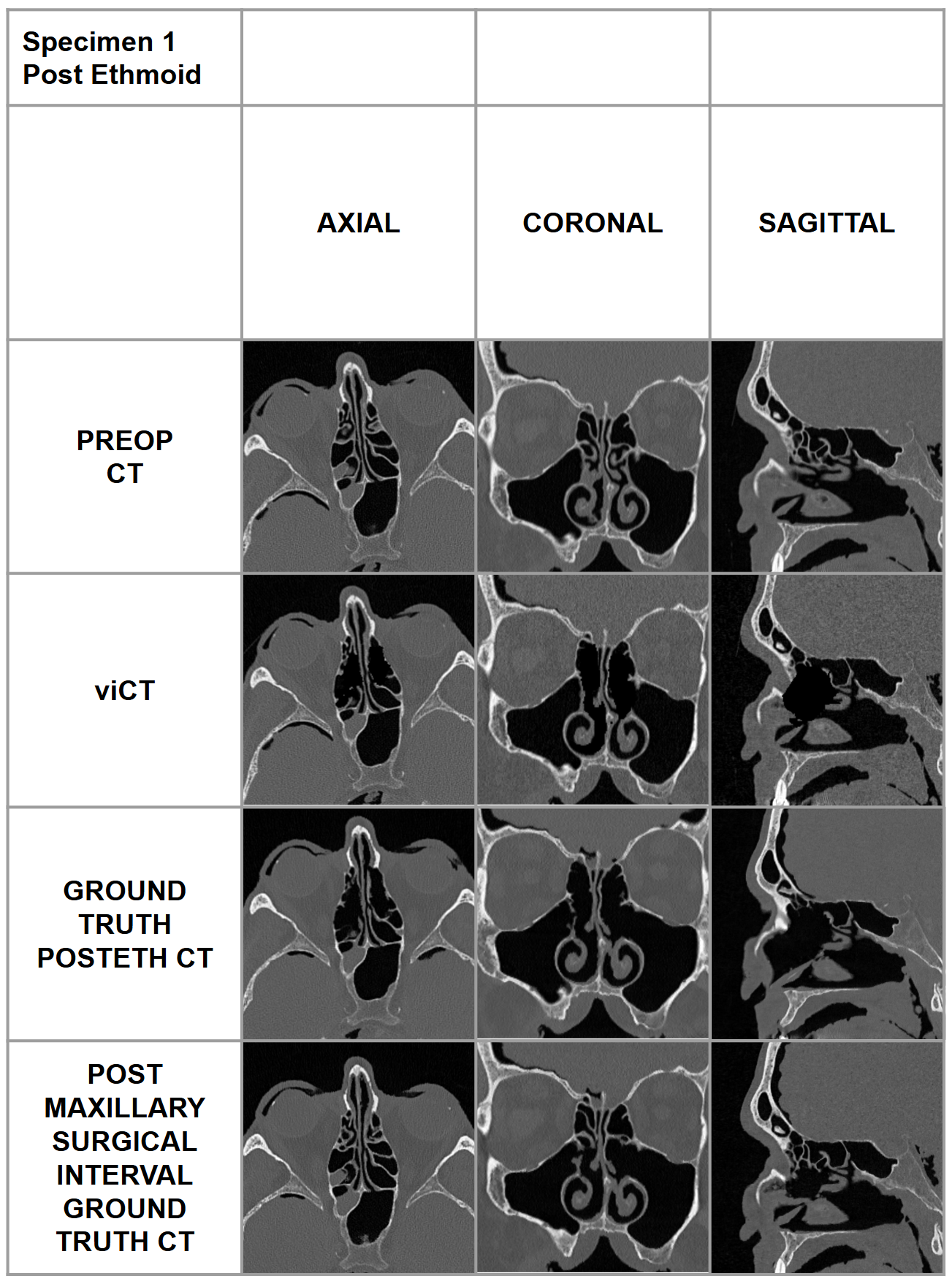}
        \caption{Post Ethmoid surgical interval}
        \label{fig:fig4B}
    \end{subfigure}
\end{figure}

\clearpage
\begin{figure}[p]\ContinuedFloat
    \centering
    \begin{subfigure}{0.9\linewidth}
        \centering
        \includegraphics[width=\linewidth]{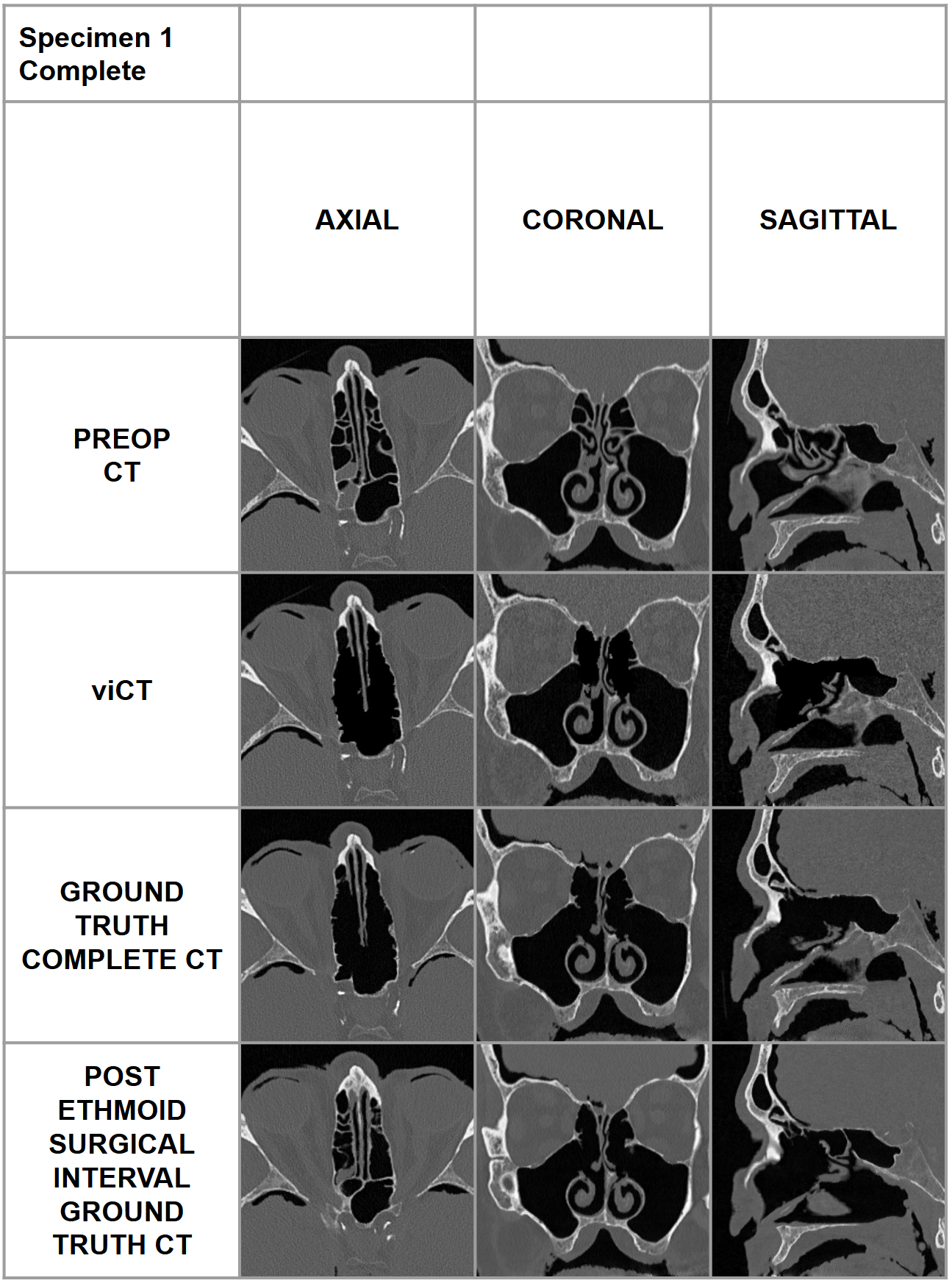}
        \caption{Complete surgical interval}
        \label{fig:fig4C}
    \end{subfigure}

    \caption{Qualitative comparison of viCT updates with ground-truth postoperative CT across surgical intervals.
    (A) Post Maxillary, (B) Post Ethmoid, and (C) Complete. Axial, coronal, and sagittal views demonstrate close correspondence between updated viCT volumes and postoperative ground-truth anatomy.}
    \label{fig:figure4}
\end{figure}
\clearpage

\begin{figure}[htp]
    \centering
    \includegraphics[width=0.9\linewidth]{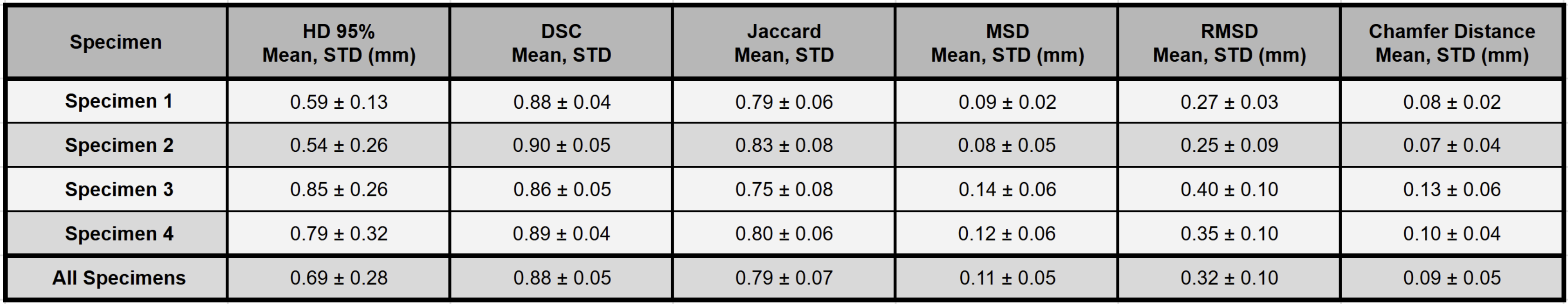} 
    \caption{Summarized quantitative data across the four cadaveric specimens.}
    \label{figure6}
\end{figure}

\subsection{Volumetric Similarity}

Updated 3D reconstructions were generated from monocular endoscopic video, coregistered to ground-truth postoperative CT, and compared using the Dice Similarity Coefficient (DSC) and Jaccard Similarity Index to quantify volumetric overlap between reconstructed and ground-truth cavity segmentations.

Volumetric agreement was consistently high across specimens. Mean DSC was $0.88 \pm 0.05$ overall ($0.88 \pm 0.04$ specimen 1, $0.90 \pm 0.05$ specimen 2, $0.86 \pm 0.05$ specimen 3, $0.89 \pm 0.04$ specimen 4; Figure 4), with bilateral side and surgical-interval stratification shown in Figure A5. Jaccard values showed similar agreement, averaging $0.79 \pm 0.07$ overall ($0.79 \pm 0.06$, $0.83 \pm 0.08$, $0.75 \pm 0.08$, $0.80 \pm 0.06$ for specimens 1--4, respectively; Figure 4; stratified results in Figure A5). Across all metrics, between-specimen variability was small (ranges of means: HD95 $0.31$ mm, DSC $0.04$, Jaccard $0.08$, MSD $0.12$ mm, RMSD $0.15$ mm, Chamfer $0.03$ mm), indicating close agreement with ground-truth cavity volumes.

\subsection{Surface Geometry Comparison}

Surface correspondence between reconstructed and ground-truth segmentations was quantified using symmetric Chamfer distance, Hausdorff Distance (HD) 95\%, mean surface distance (MSD), and root-mean-square surface distance (RMSD), providing complementary measures of global alignment and local surface fidelity relative to postoperative CT.

Geometric agreement was strong and consistent across all four cadaveric specimens. Mean HD95 was $0.69 \pm 0.28$ mm overall ($0.59 \pm 0.13$, $0.54 \pm 0.26$, $0.85 \pm 0.26$, and $0.79 \pm 0.32$ mm for specimens 1–4; Figure 4, stratified in Figure A5). Mean Chamfer distance was $0.09 \pm 0.05$ mm overall ($0.08 \pm 0.02$, $0.07 \pm 0.04$, $0.13 \pm 0.06$, $0.10 \pm 0.04$ mm), indicating close point-wise surface correspondence.

MSD averaged $0.11 \pm 0.05$ mm across specimens ($0.09 \pm 0.02$, $0.08 \pm 0.05$, $0.14 \pm 0.06$, $0.12 \pm 0.06$ mm), and RMSD averaged $0.32 \pm 0.10$ mm ($0.27 \pm 0.03$, $0.25 \pm 0.09$, $0.40 \pm 0.10$, $0.35 \pm 0.10$ mm), demonstrating small and evenly distributed surface deviations. Minor outliers were observed but remained within expected anatomical and reconstruction variability. Collectively, these metrics indicate minimal surface error relative to ground truth and support the accuracy and robustness of the reconstruction pipeline across four cadaveric specimens and surgical intervals.

\section{Discussion}\label{sec5}

We present a method for iterative updating of preoperative CT (pCT) using intraoperative 3D reconstructions, demonstrating high preliminary agreement with ground-truth cross-sectional imaging. A depth-supervised NeRF with simulated stereo provides metrically scaled reconstruction of the surgical field without external tracking hardware. The viCT pipeline maps both pCT and reconstruction into the native CT grid and performs ray-based occupancy comparison to remove resected voxels and update boundaries, producing a standard CT-format volume viewable in axial, coronal, and sagittal planes.

Despite persistent rates of incomplete resection and revision ESS, few systems update imaging intraoperatively to reflect evolving anatomy. Existing IGS platforms (e.g., Acclarent TruDi, Stryker TGS) rely on static pCT overlays. TruDi’s Fast Anatomic Mapping requires manual probe-based registration, interrupting workflow and limiting spatial completeness.

Vision-based updating has been explored, including monocular depth with TSDF fusion for surface evolution \cite{Mangulabnan2024EndoscopicChisel}; however, reported errors may limit fine-scale ESS applicability, and CT-format volumes are not generated. While Gaussian Splatting enables efficient rendering, its view-dependent representation is appearance-driven. In contrast, our depth-supervised NeRF with virtual stereo enforces geometric consistency and metric scale, enabling dense, voxel-level CT updating. \cite{Liu2025Survey3DReconstructionNeRF3DGS}

Ongoing work targets automation and acceleration of reconstruction, registration, and updating. Reconstruction currently requires <10 minutes and CT updating <2 minutes (dual RTX 5080 GPUs), though landmark-based registration remains a bottleneck. Given ESS durations of 1–4 hours and intraoperative CT delays of ~30 minutes, viCT could be computed in parallel when iCT would otherwise be considered but impractical. Future work will incorporate Dense Correspondence Analysis (DeCA) \cite{Rolfe2023DeCA} with affine and nonlinear refinement using ANTs \cite{Avants2009ANTS}, alongside validation on live surgical video to assess robustness under realistic intraoperative conditions.

\section{Conclusion}\label{sec6}

This work demonstrates the feasibility of sequential CT updating from monocular endoscopy in a cadaveric ESS model. By integrating metrically scaled 3D reconstructions from depth-supervised NeRF with the original CT via voxel-level, ray-based occupancy comparison, the system generates CT-format updates that preserve anatomical context while reflecting surgical progression.

Across four cadaveric specimens and four surgical stages, viCT achieved submillimeter accuracy, with mean HD95 $0.69 \pm 0.28$ mm, DSC $0.88 \pm 0.05$, Jaccard $0.79 \pm 0.07$, MSD $0.11 \pm 0.05$ mm, RMSD $0.32 \pm 0.10$ mm, and Chamfer $0.09 \pm 0.05$ mm.

Unlike existing IGS or prior vision-based fusion approaches, viCT provides volumetric CT visualization of tissue removal without external tracking hardware, offering an intuitive and quantitative representation of evolving resection boundaries.

Future work will focus on automated registration, validation in live surgical cases, and computational optimization for intraoperative deployment, alongside formal evaluation of clinical utility and workflow integration.

\backmatter

\section*{Statements and Declarations}\label{sec7}

\subsection*{Funding}

This work was supported by a Phase I grant from the Washington Research Foundation (WRF) under the project \emph{“Next Generation Image Guided Endoscopic Sinus and Skull-base Surgery”}, awarded to the University of Washington for the period September 2025–September 2026, and a NIH R25 grant awarded to Graham Harris at the University of Washington Medical Center for the period July 2025-July 2026.

\subsection*{Competing Interests}

Dr. Bly is a cofounder and holds a financial interest of ownership equity with Wavely Diagnostics Inc and Apertur Inc. He is a consultant and stockholder of Spiway LLC. These are not related to this study. All other authors do not have information to disclose. 

\subsection*{Ethics Approval}

This study did not involve living human participants. All experiments were conducted exclusively on human cadaveric specimens and associated de-identified imaging data. All cadaveric material was obtained and handled in accordance with institutional policies governing the ethical use of human anatomical specimens.

\subsection*{Consent to participate}

No human data or images from living individuals are included in this article.

\subsection*{Data and Code Availability}

The data and code used during this study are not publicly available due to ongoing intellectual property protection related to a pending patent application, but are available from the corresponding author upon reasonable request and subject to institutional restrictions.

\subsection*{Author Contributions}

Conceptualization: Nicole M. Gunderson; Jeremy S. Ruthberg;  Randall A. Bly; Waleed M. Abuzeid; Eric J. Seibel. Methodology: Nicole M. Gunderson; Pengcheng Chen. Formal analysis and investigation: Nicole M. Gunderson; Graham J. Harris; Jeremy S. Ruthberg; Di Mao. Writing - original draft preparation: Nicole M. Gunderson. Writing - review and editing: Nicole M. Gunderson; Graham J. Harris; Jeremy S. Ruthberg; Pengcheng Chen; Di Mao; Randall A. Bly; Waleed M. Abuzeid; Eric J. Seibel. Funding acquisition: Nicole M. Gunderson; Jeremy S. Ruthberg. Resources: Randall A. Bly; Waleed M. Abuzeid; Eric J. Seibel. Supervision: Randall A. Bly; Waleed M. Abuzeid; Eric J. Seibel.

\bibliography{sn-bibliography}% common bib file

\begin{appendices}

\section{Supplementary Figures}\label{secA1}

% Second figure: quantitative data
\begin{figure}[p]
    \centering
    \includegraphics[width=\textwidth, height=0.9\textheight, keepaspectratio]{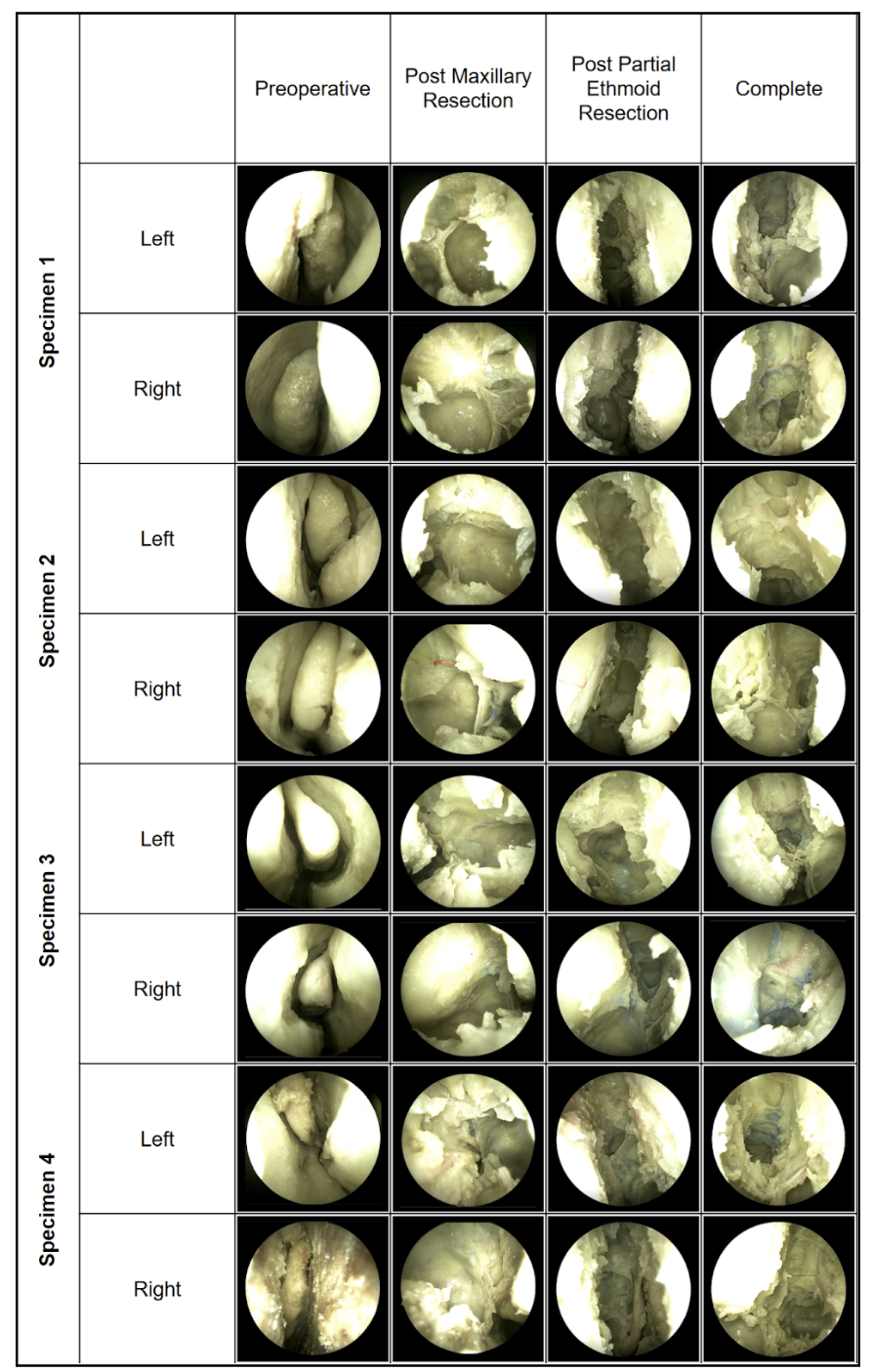}
    \caption{Sample monocular endoscopic views from videos used as input to the 3D reconstruction and viCT algorithms.}
    \label{fig:figure3}
\end{figure}

\begin{figure}[p]
    \centering
    \begin{subfigure}{0.9\linewidth}
        \centering
        \includegraphics[width=\linewidth]{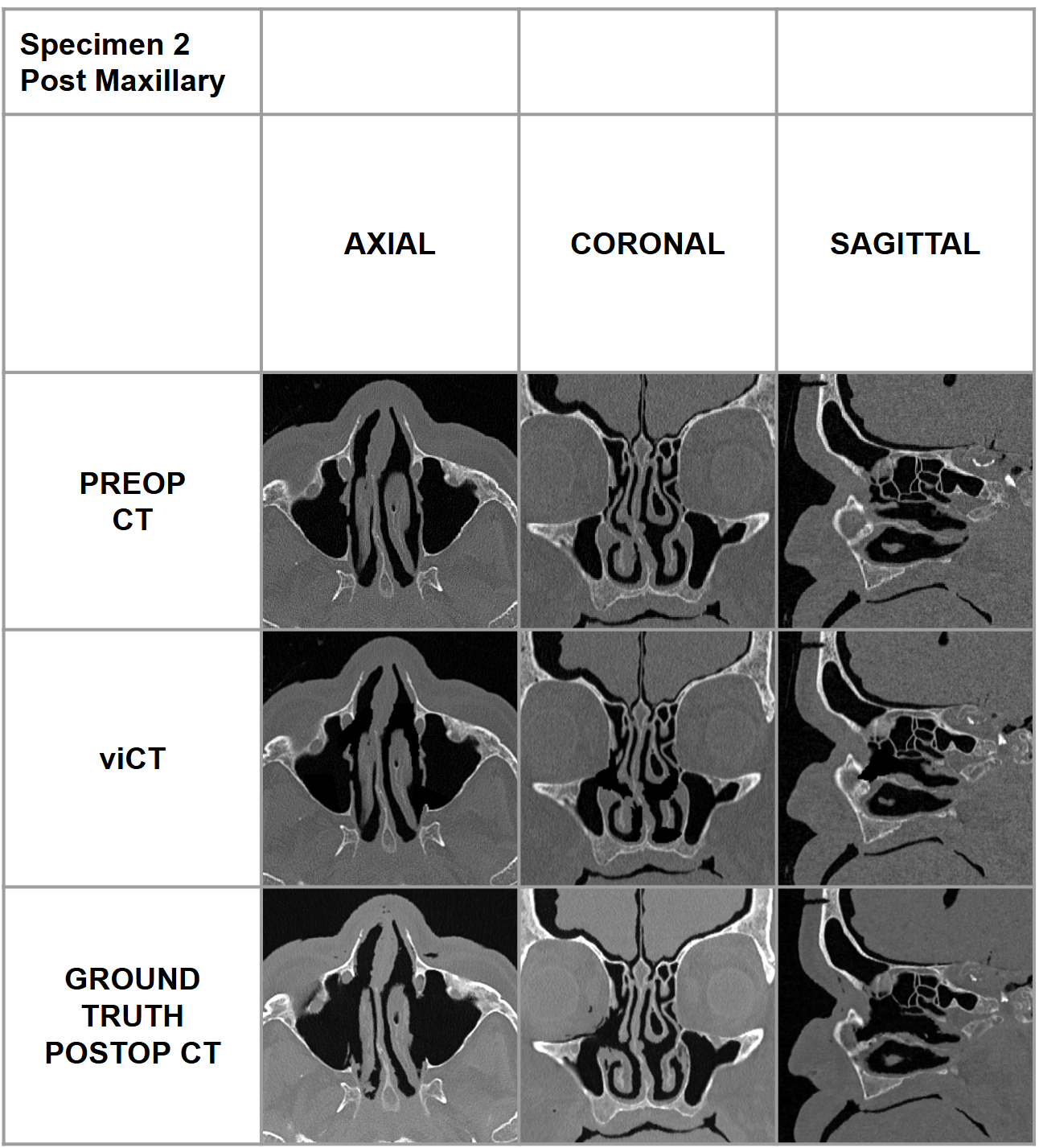}
        \caption{Post Maxillary surgical interval}
        \label{fig:fig7A}
    \end{subfigure}
\end{figure}

\begin{figure}[p]\ContinuedFloat
    \centering
    \begin{subfigure}{0.9\linewidth}
        \centering
        \includegraphics[width=\linewidth]{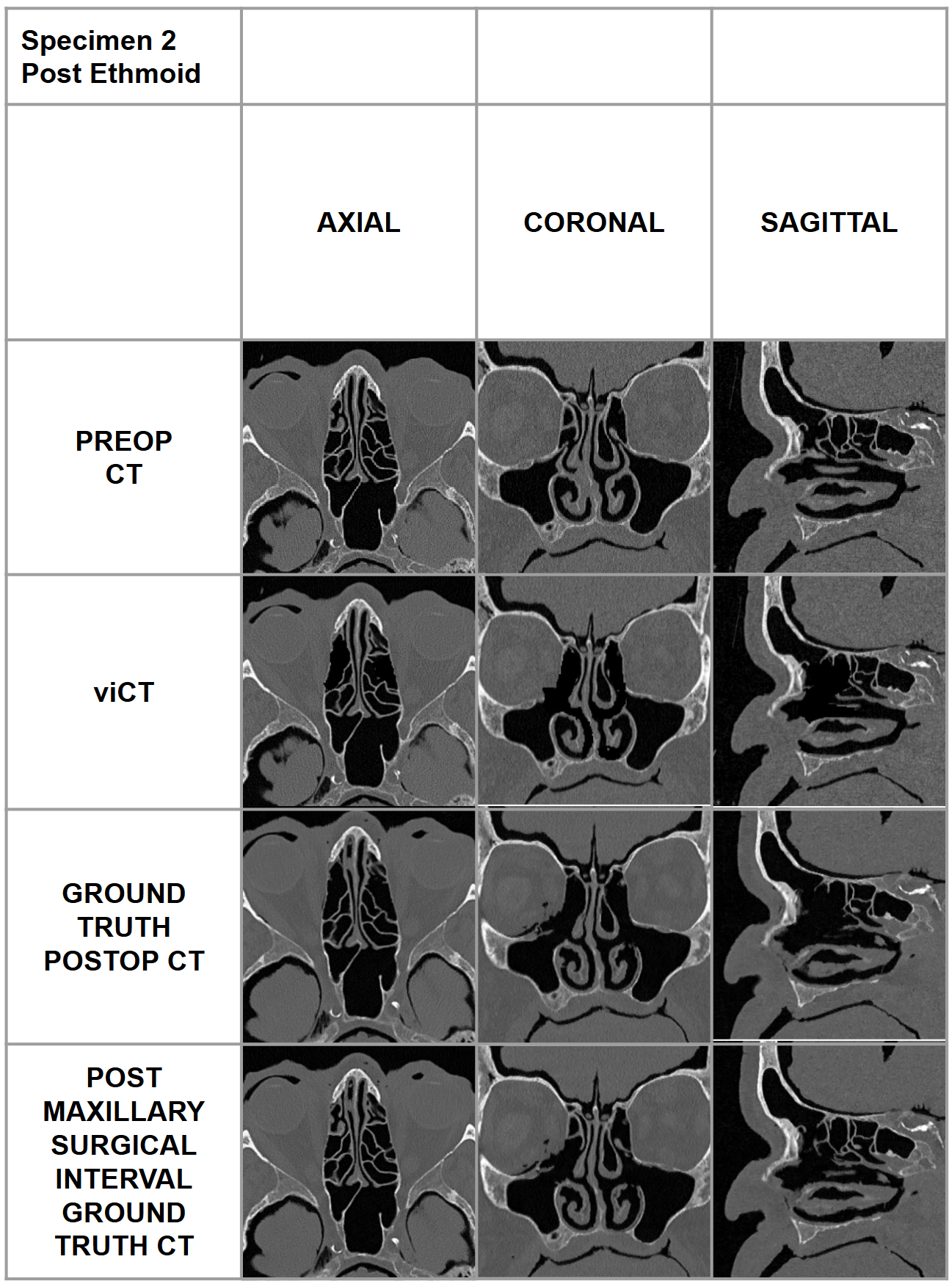}
        \caption{Post Ethmoid surgical interval}
        \label{fig:fig7B}
    \end{subfigure}
\end{figure}

\begin{figure}[p]\ContinuedFloat
    \centering
    \begin{subfigure}{0.9\linewidth}
        \centering
        \includegraphics[width=\linewidth]{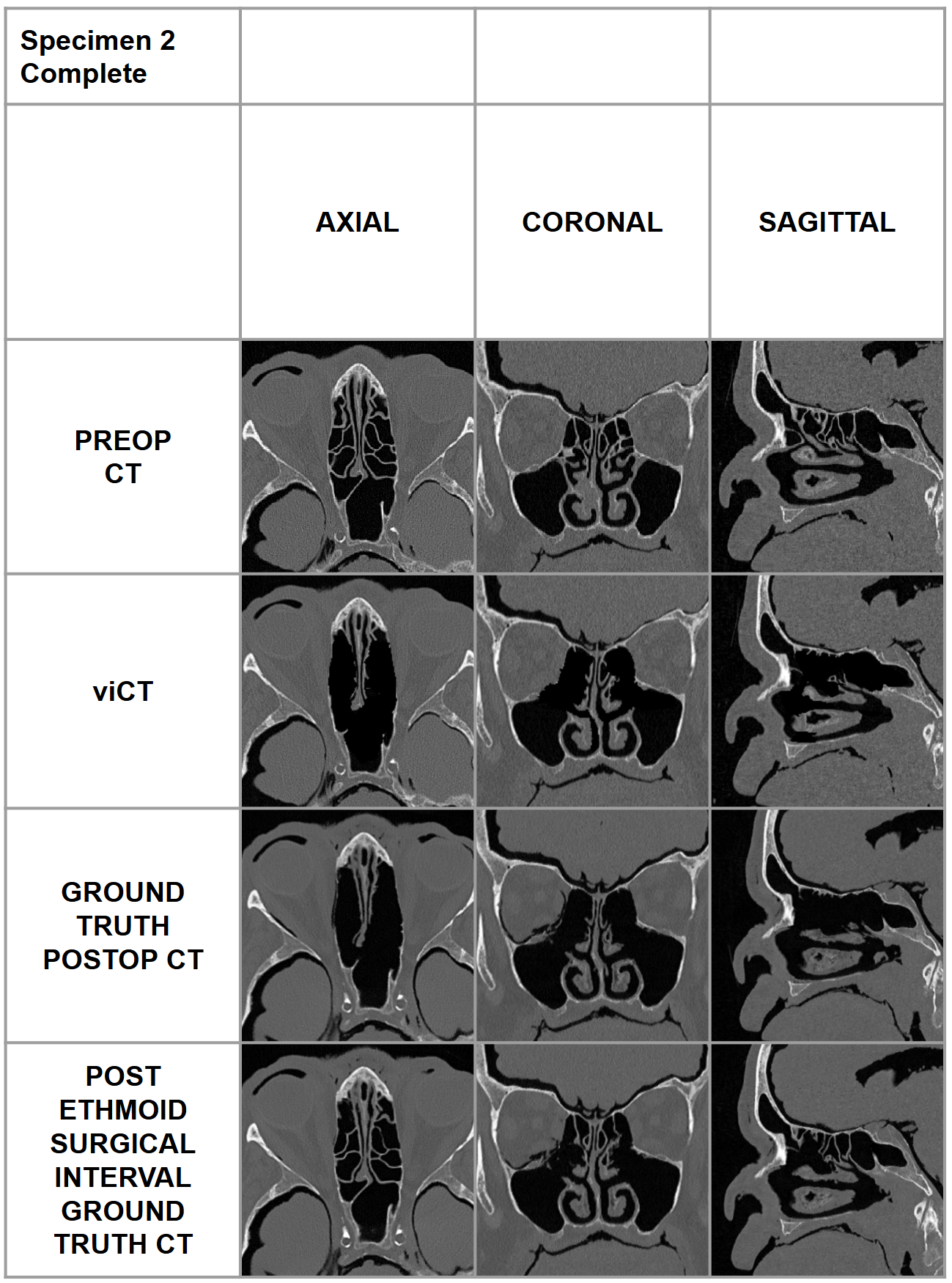}
        \caption{Complete surgical interval}
        \label{fig:fig7C}
    \end{subfigure}

    \caption{Qualitative comparison of viCT updates with ground-truth postoperative CT across surgical intervals.
    (A) Post Maxillary, (B) Post Ethmoid, and (C) Complete. Axial, coronal, and sagittal views demonstrate close correspondence between updated viCT volumes and postoperative ground-truth anatomy.}
    \label{fig:figure7}
\end{figure}

\begin{figure}[p]
    \centering
    \begin{subfigure}{0.9\linewidth}
        \centering
        \includegraphics[width=\linewidth]{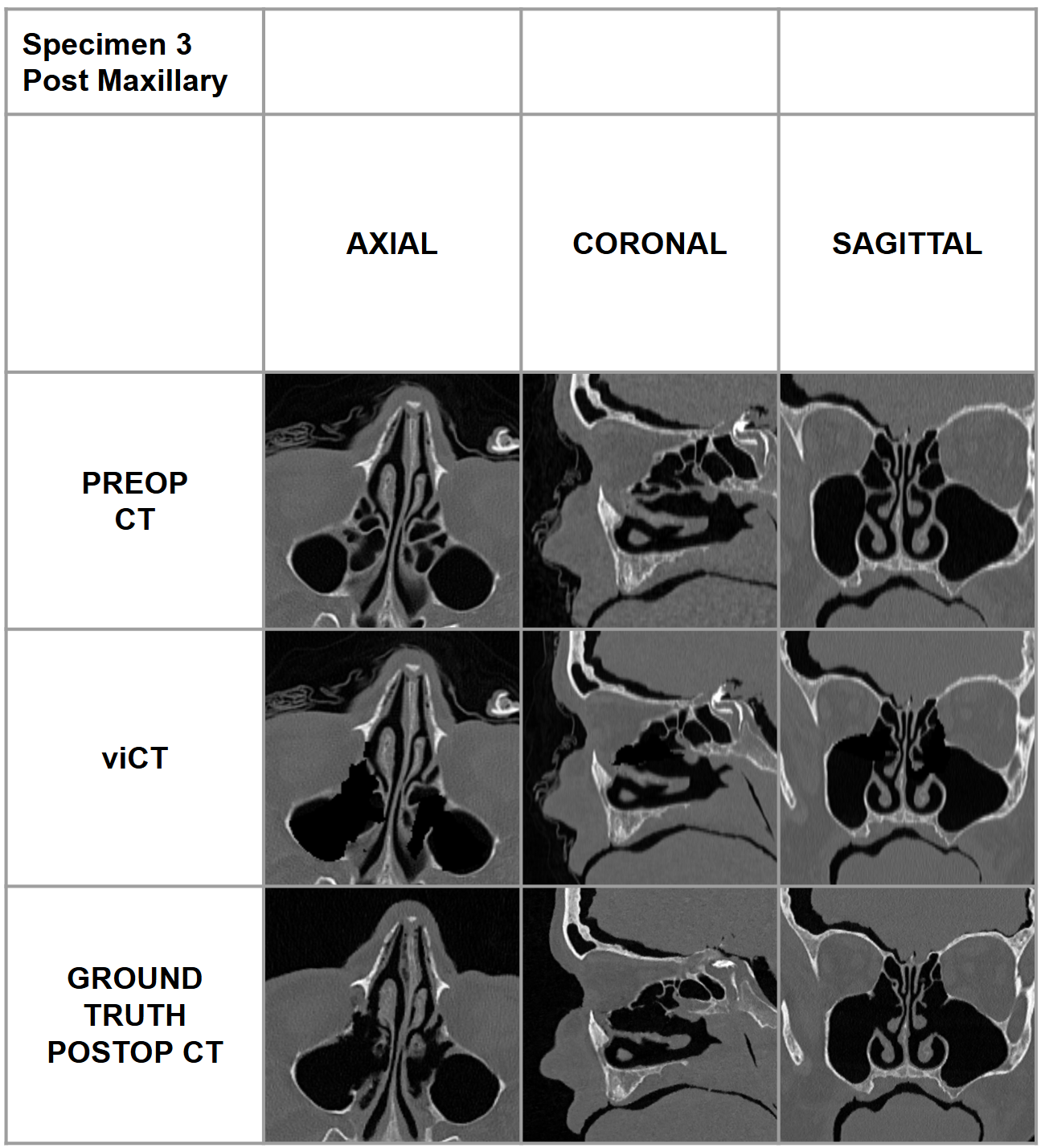}
        \caption{Post Maxillary surgical interval}
        \label{fig:fig8A}
    \end{subfigure}
\end{figure}

\begin{figure}[p]\ContinuedFloat
    \centering
    \begin{subfigure}{0.9\linewidth}
        \centering
        \includegraphics[width=\linewidth]{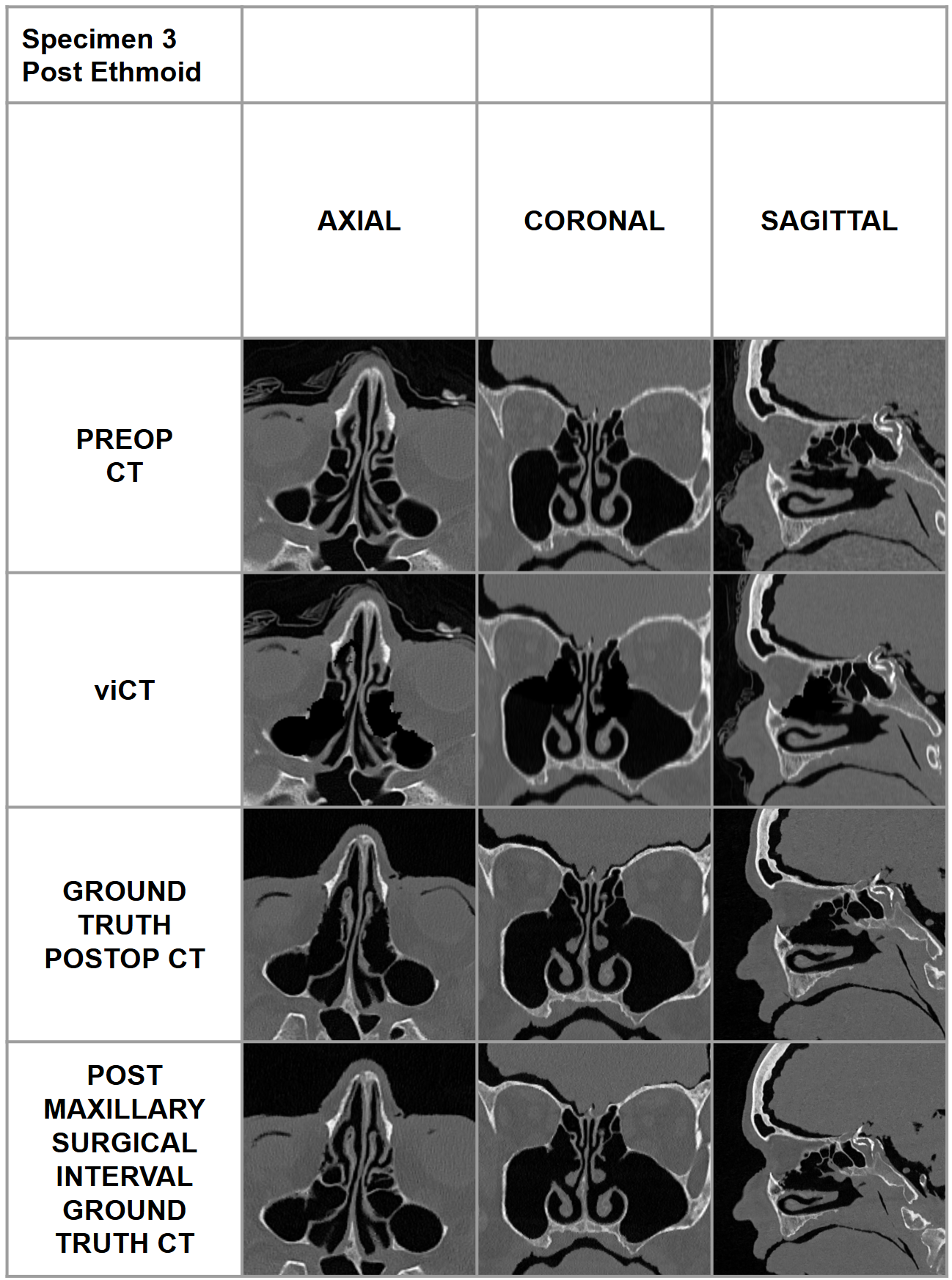}
        \caption{Post Ethmoid surgical interval}
        \label{fig:fig8B}
    \end{subfigure}
\end{figure}

\begin{figure}[p]\ContinuedFloat
    \centering
    \begin{subfigure}{0.9\linewidth}
        \centering
        \includegraphics[width=\linewidth]{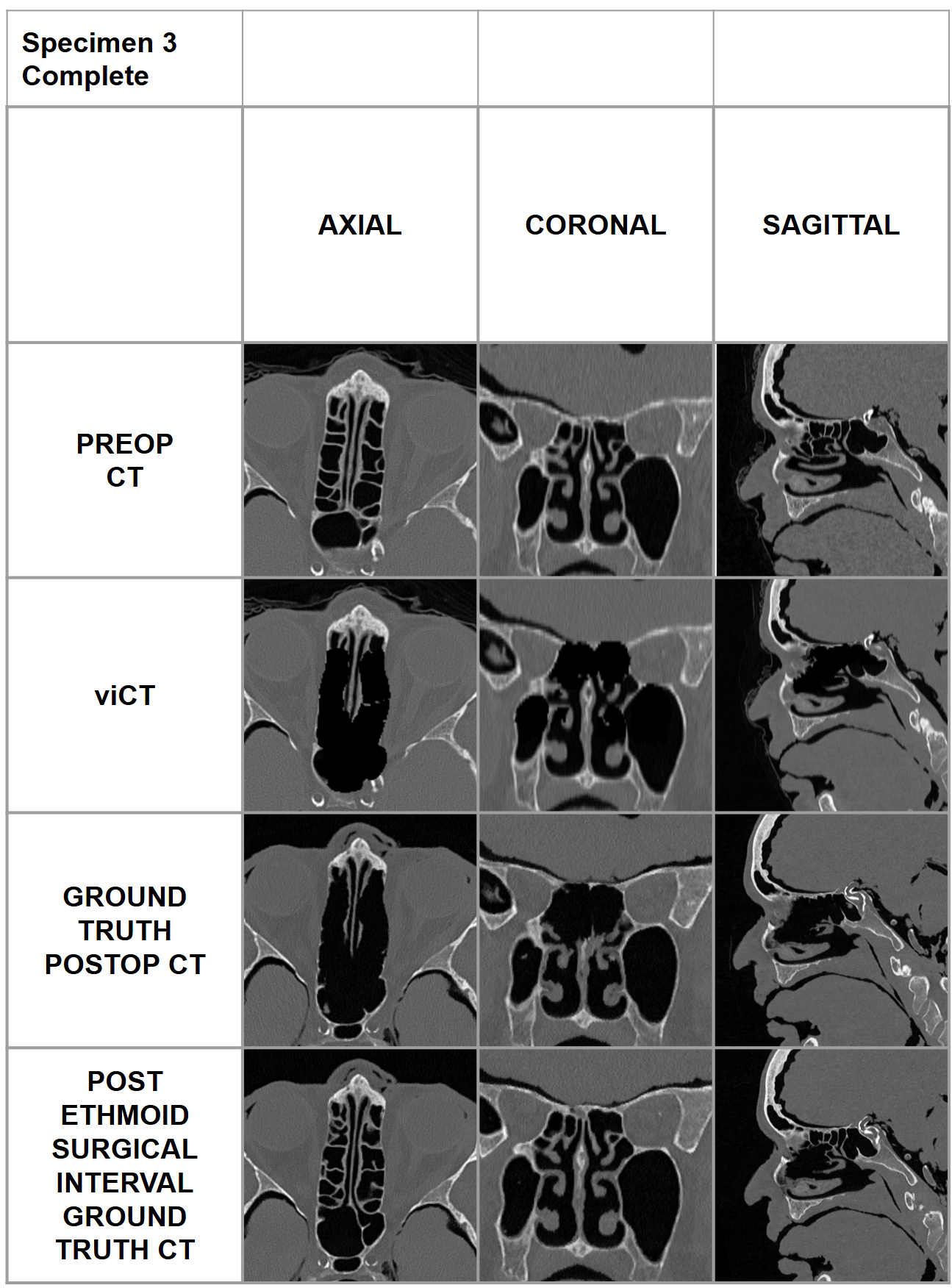}
        \caption{Complete surgical interval}
        \label{fig:fig8C}
    \end{subfigure}

    \caption{Qualitative comparison of viCT updates with ground-truth postoperative CT across surgical intervals.
    (A) Post Maxillary, (B) Post Ethmoid, and (C) Complete. Axial, coronal, and sagittal views demonstrate close correspondence between updated viCT volumes and postoperative ground-truth anatomy.}
    \label{fig:figure8}
\end{figure}

\begin{figure}[p]
    \centering
    \begin{subfigure}{0.9\linewidth}
        \centering
        \includegraphics[width=\linewidth]{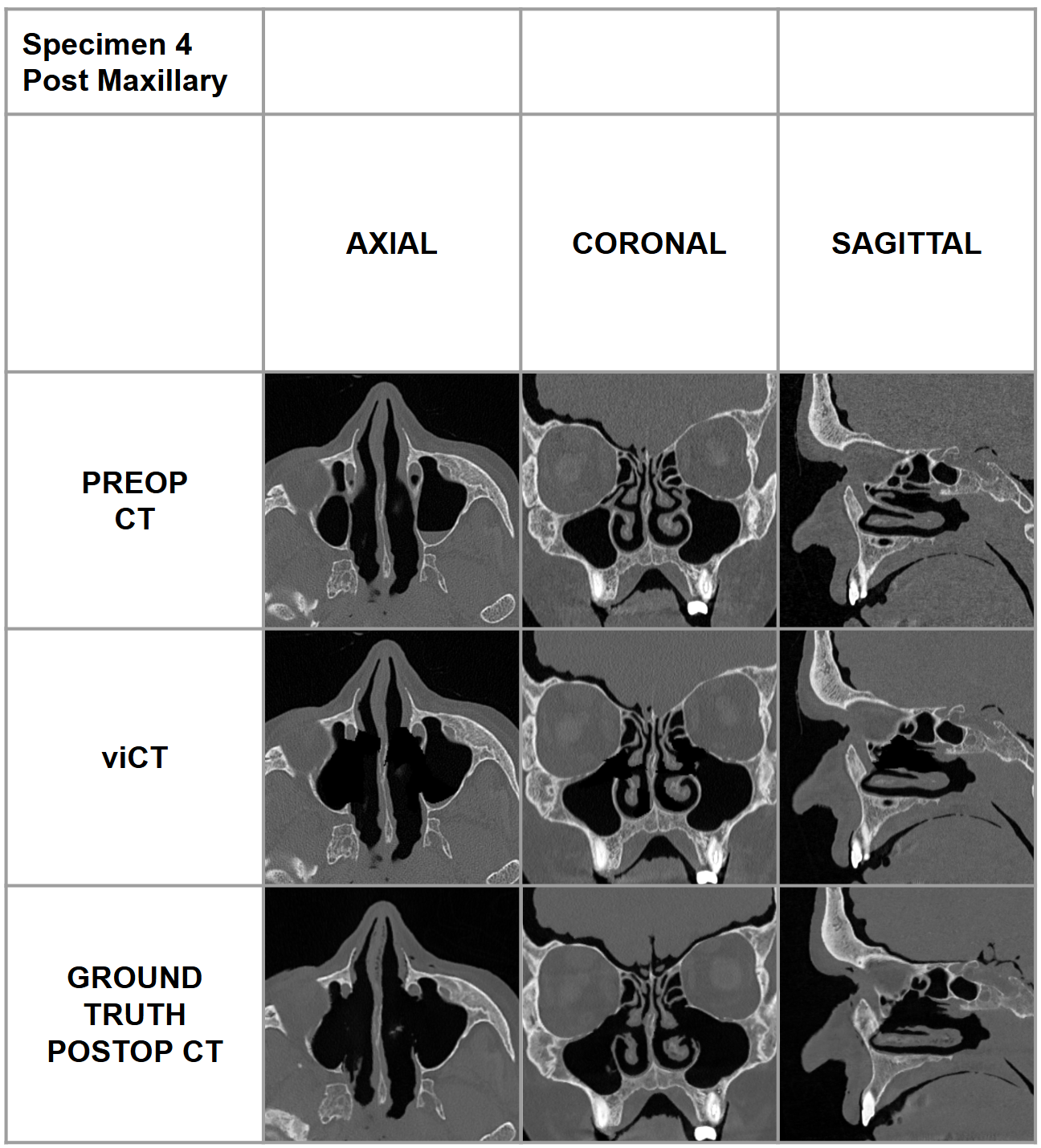}
        \caption{Post Maxillary surgical interval}
        \label{fig:fig9A}
    \end{subfigure}
\end{figure}

\begin{figure}[p]\ContinuedFloat
    \centering
    \begin{subfigure}{0.9\linewidth}
        \centering
        \includegraphics[width=\linewidth]{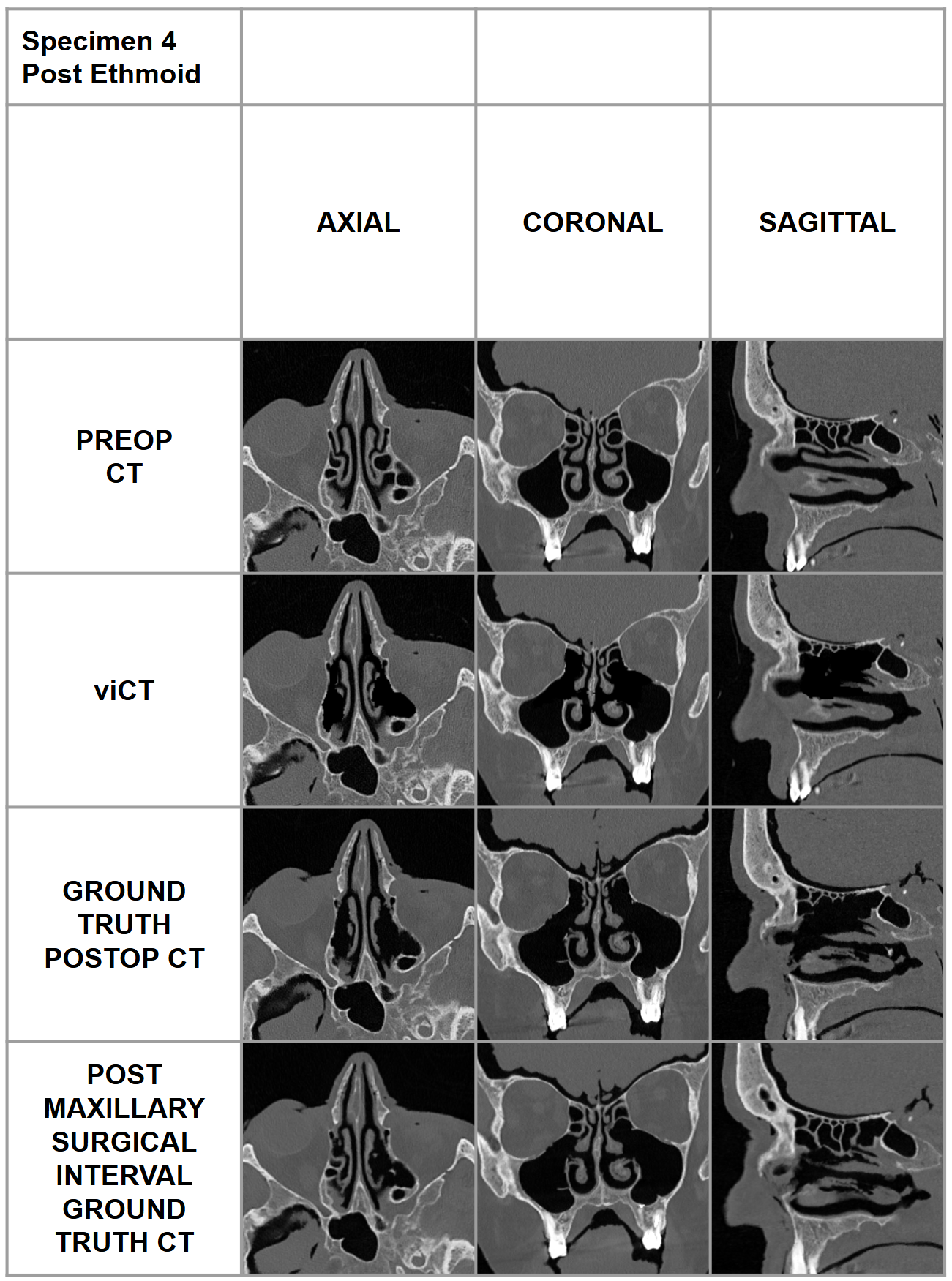}
        \caption{Post Ethmoid surgical interval}
        \label{fig:fig9B}
    \end{subfigure}
\end{figure}

\begin{figure}[p]\ContinuedFloat
    \centering
    \begin{subfigure}{0.9\linewidth}
        \centering
        \includegraphics[width=\linewidth]{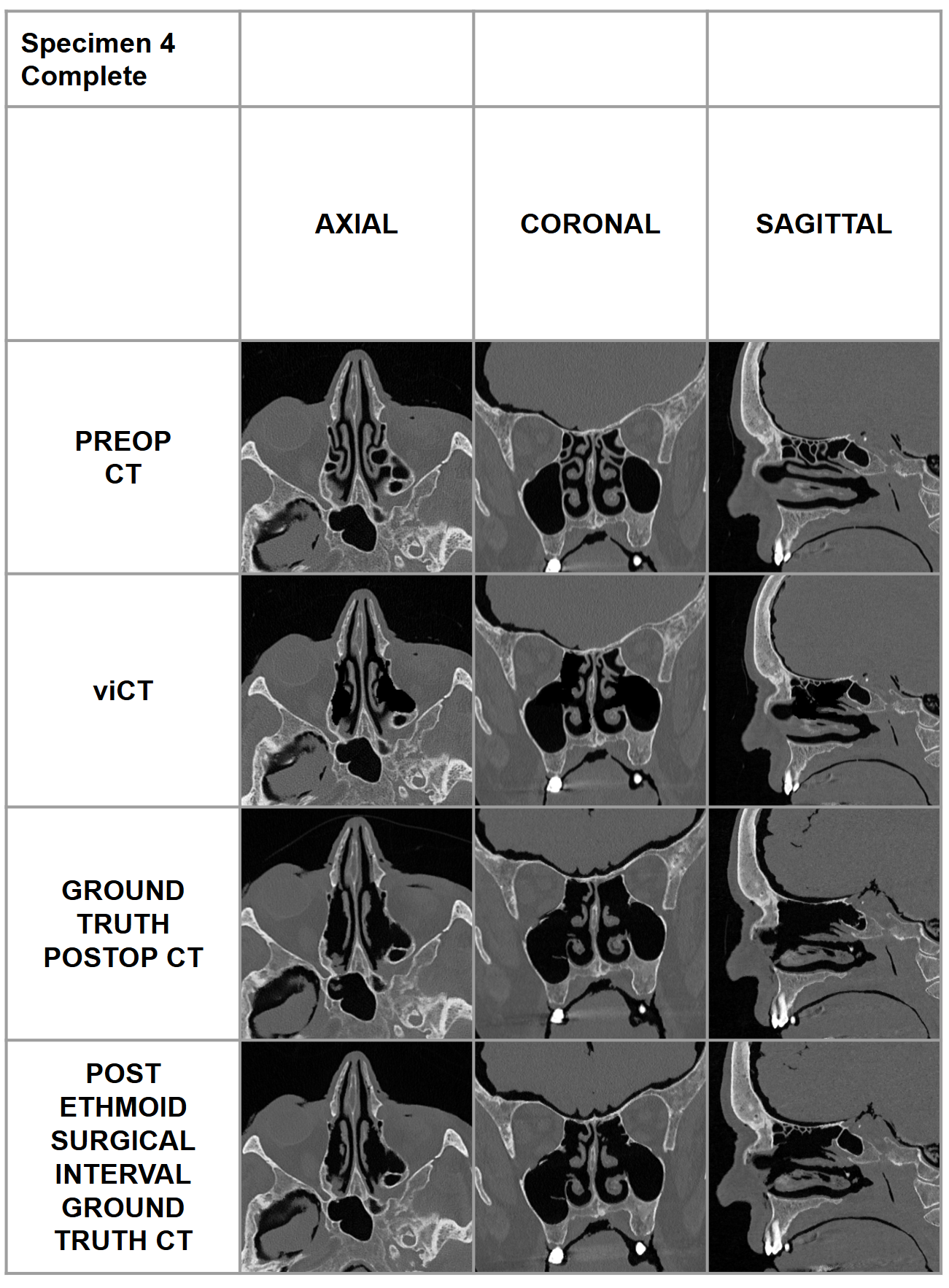}
        \caption{Complete surgical interval}
        \label{fig:fig9C}
    \end{subfigure}

    \caption{Qualitative comparison of viCT updates with ground-truth postoperative CT across surgical intervals.
    (A) Post Maxillary, (B) Post Ethmoid, and (C) Complete. Axial, coronal, and sagittal views demonstrate close correspondence between updated viCT volumes and postoperative ground-truth anatomy.}
    \label{fig:figure9}
\end{figure}

% Second figure: quantitative data
\begin{figure}[p]
    \centering
    \includegraphics[width=\textwidth, height=0.9\textheight, keepaspectratio]{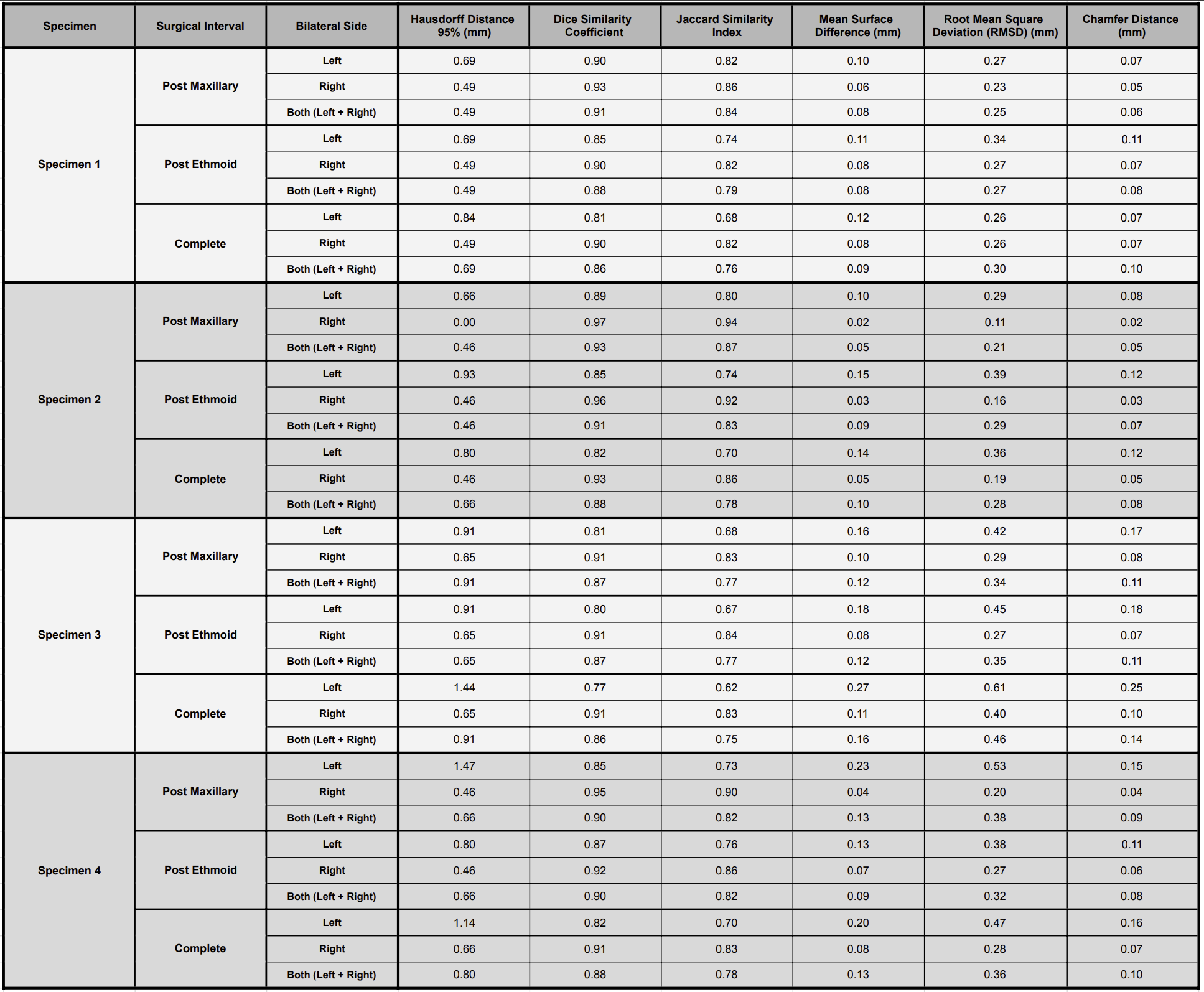}
    \caption{Full quantitative data across the four cadaveric specimens.}
    \label{fig:figure5}
\end{figure}

\end{appendices}

\end{document}